\documentclass[letterpaper]{article} 
\usepackage{aaai2026}  
\usepackage{times}  
\usepackage{helvet}  
\usepackage{courier}  
\usepackage[hyphens]{url}  
\usepackage{graphicx} 
\usepackage{natbib}  
\usepackage{caption} 
\usepackage{algorithm}
\usepackage{algorithmic}
\usepackage{kotex}
\usepackage{amsfonts}
\usepackage{placeins}
\usepackage{url}

\usepackage{newfloat}
\usepackage{listings}
\usepackage{multirow}

\DeclareCaptionStyle{ruled}{labelfont=normalfont,labelsep=colon,strut=off} 
\floatstyle{ruled}
\newfloat{listing}{tb}{lst}{}
\floatname{listing}{Listing}

\title{Prompt the Unseen: Evaluating Visual-Language Alignment Beyond Supervision}

\author{
    Raehyuk Jung\textsuperscript{\rm 1},  
    Seungjun Yu\textsuperscript{\rm 1},  
    Hyunjung Shim\textsuperscript{\rm 1}
}
\affiliations{
    \textsuperscript{\rm 1}Computer Vision and Machine Learning (CVML), KAIST \\
    Seocho-gu, Seoul, South Korea \\
    mulnyangi@naver.com, seungjunyu@kaist.ac.kr, kateshim@kaist.ac.kr
}

\usepackage{bibentry}

\nocopyright

\begin{document}

\maketitle

\renewcommand{\thefootnote}{\fnsymbol{footnote}}
\footnotetext[1]{\textbf{Code available at: }
\mbox{\url{https://github.com/raemoi93/PromptTheUnseen}}}
\renewcommand{\thefootnote}{\arabic{footnote}}

\begin{abstract}
Vision-Language Models (VLMs) combine a vision encoder and a large language model (LLM) through alignment training, showing strong performance on multimodal tasks. A central component in this architecture is the projection layer, which maps visual features into the LLM’s embedding space. Despite its importance, its ability to generalize to unseen visual concepts has not been systematically evaluated. 
To address this, we propose a benchmark for evaluating projection layer generalization. We adapt object detection datasets—rich in fine-grained annotations—into a prompting format, and design train/test splits with disjoint label sets. This setup enables precise control over seen and unseen concept separation. 
Experimental results show that the projection layer retains over 79-88\% of the performance on unseen classes compared to seen ones, across various settings. This suggests a non-trivial level of generalization, even without explicit alignment supervision on those concepts. 
We further analyze this behavior through a mechanistic interpretability lens. Our findings suggest that the feed-forward network in the projection layer operates like a key-value memory, processing seen and unseen tokens in similar ways. This study introduces a new evaluation framework for alignment generalization and highlights the potential for efficient VLM training with limited aligned data.

\end{abstract}

\section{Introduction}



The rise of large language models (LLMs) has transformed natural language processing, enabling strong generalization across a wide range of linguistic tasks. As these models expand to support visual inputs, Vision-Language Models (VLMs) are becoming key to building general-purpose multimodal systems. From image captioning to visual assistants, VLMs now drive many real-world applications, with models such as GPT-4V and Gemini leading this transition.



At the core of modern VLM design lies a simple yet powerful idea: instead of training large multimodal architectures from scratch, one can connect a pretrained vision encoder and a pretrained LLM using a lightweight projection layer. This approach offers two major advantages—computational efficiency and modular flexibility. It allows the system to inherit the generalization capabilities of both components while requiring only modest alignment training to bridge the modalities. This paradigm has been widely adopted in recent models such as Qwen-VL~\cite{bai2025qwen2} and InternVL-2.5~\cite{chen2024expanding}.


\begin{figure}[t]
  \centering
  \makebox[\linewidth]{%
    \includegraphics[width=1.10\linewidth]{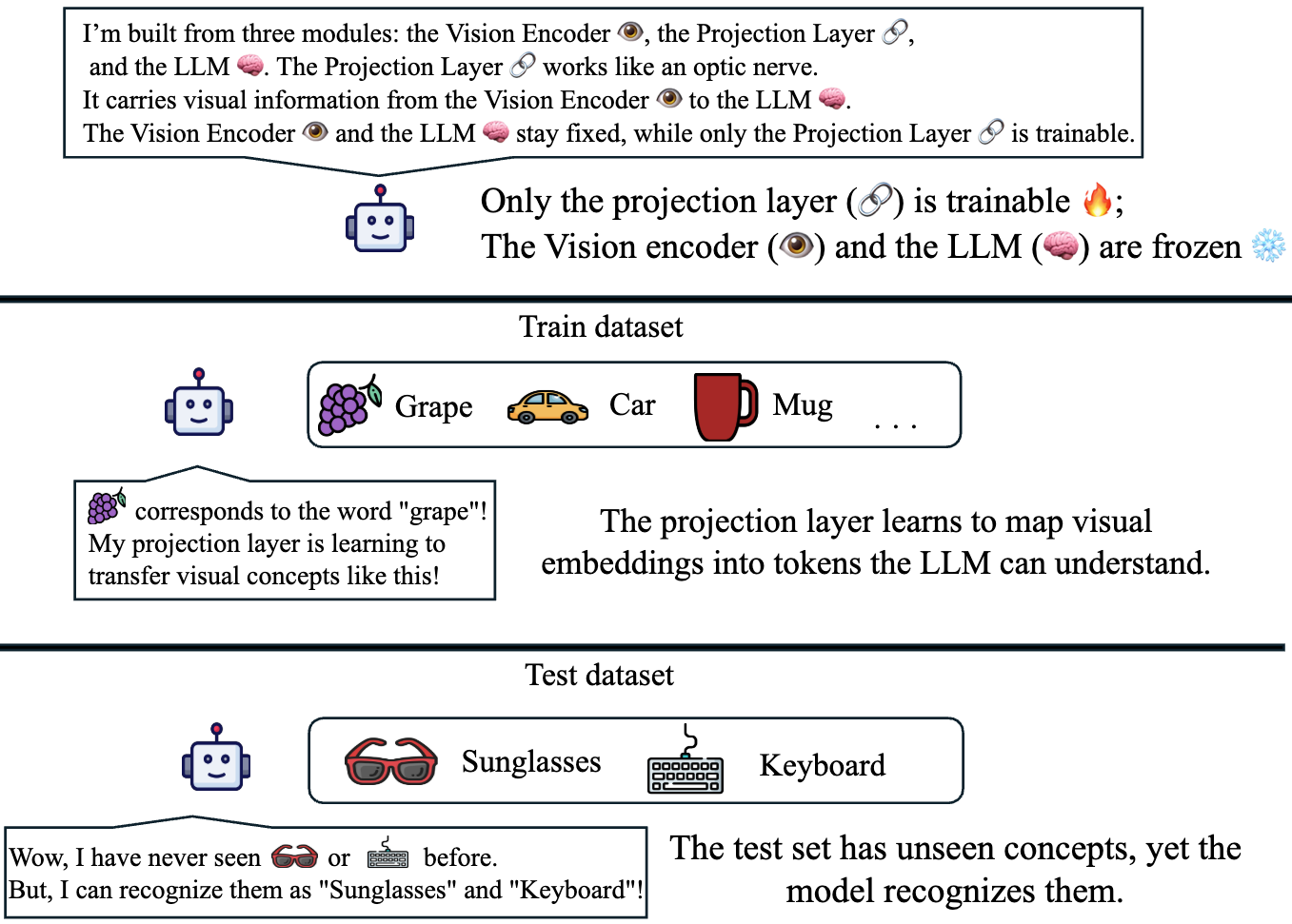}%
  }
  \caption{This figure illustrates our overall research question: \textit{Can the projection layer effectively transfer visual information of unseen concepts?} At the top, the VLM is shown as consisting of three modules: the vision encoder, the projection layer, and the LLM. The vision encoder and LLM are frozen, while only the projection layer is trainable. In the middle, we provide simplified, illustrative training examples (e.g., grape, car, mug) showing how the projection layer learns to map fixed visual embeddings into tokens that the frozen LLM can recognize. At the bottom, we depict hypothetical test examples (sunglasses, keyboard), demonstrating the model’s ability to recognize novel objects, even though the projection layer was never trained on them. These examples are schematic and meant to aid intuition, rather than drawn from the actual datasets used in our experiments.}
  \label{fig:easy_understanding}
\end{figure}

However, this modular approach introduces a critical dependency: the projection layer becomes the sole pathway through which visual information reaches the language model. Despite its functional importance, the projection layer is orders of magnitude smaller in both training data and parameter count. For example, in LLaVA~\cite{liu2023visual}, the vision encoder (CLIP-Large~\cite{radford2021learning}) is trained on 400 million image-text pairs, and the LLM (Vicuna-13B~\cite{chiang2023vicuna}) is trained on over two trillion tokens. The projection layer connecting the two, by contrast, is trained on only 750,000 alignment examples and consists of just 4.2 million parameters. This sharp asymmetry raises an important but underexplored question:
Can such a small module generalize to visual concepts it has never seen during alignment training? As illustrated in figure~\ref{fig:easy_understanding}, we present this research question with a visual aid that clarifies our setting and motivation.

This question is critical in scalable and efficient VLM design. If the projection layer fails to generalize, one must rely on increasingly large and diverse alignment datasets, which would erode the efficiency and modularity benefits of current VLM architectures. Although projection-based models report high overall performance as reported in various studies~\cite{bai2025qwen2, li2024llava}, they are typically trained and evaluated on datasets that lack structured annotations distinguishing seen from unseen visual concepts, making it difficult to draw conclusions about generalization behavior.

To properly assess this generalizibility in unseen categories, a carefully curated benchmark is necessary where a clear separation between seen and unseen visual concepts is guaranteed, providing controlled, prompt-based evaluation. However, no such dataset currently exists. Large-scale multimodal datasets such as LAION~\cite{schuhmann2022laion} offer scale, but lack structural annotation to isolate concept-level generalization to construct a suituable test set with the desired properties.

\begin{figure}[t]
  \centering
  \includegraphics[width=\linewidth]{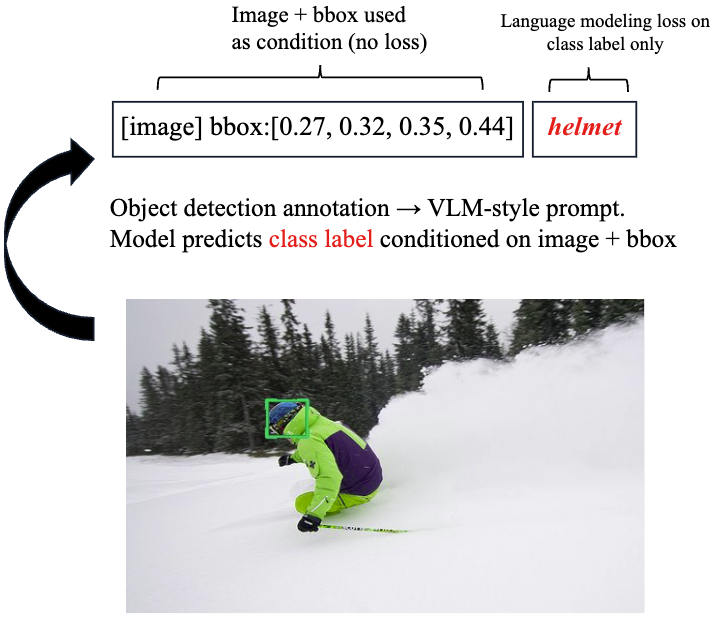}
  \caption{
 This figure illustrates how an object detection annotation is converted into a VLM-style prompt. Each annotation consists of an image, bounding box coordinates, and a class label. We convert these into a prompt where the image, bounding box, and class label appear sequentially. The model is trained to predict the class label conditioned on the image and bounding box, with the language modeling loss applied only to the class label. For clarity, we show a hypothetical bounding box and object in the exemplar image, but such boxes are not explicitly drawn during training or evaluation.}
  \label{fig:conversion_example}
\end{figure}

In this work, we propose a new evaluation framework to address this gap. Our key idea is to leverage object detection datasets—particularly Visual Genome~\cite{krishna2017visual}, which provides the largest number of fine-grained unique class labels among commonly used datasets and includes explicit bounding boxes and class labels. Unlike classification datasets, detection datasets allow multiple labeled regions per image and mitigate label ambiguity often found in image-level annotations~\cite{yun2021re}. We partition the label space into disjoint seen and unseen sets, and convert bounding box annotations into prompt-style training samples to construct a vision-language alignment dataset with fine-grained control over visual concepts. Figure~\ref{fig:conversion_example} illustrates an exemplar annotation and its conversion into a VLM-style prompt. We then train only the projection layer while keeping the vision encoder and LLM frozen, ensuring that any observed generalization arises solely from the alignment process.

To evaluate the resulting models, we first design a multiple-choice QA benchmark with separate splits for seen and unseen labels. We design a series of experiments to rigorously evaluate the model’s generalization capability across diverse settings, including challenging scenarios where the language model is as small as 0.6B. Even under limited capacity, the model achieves over 79\%—and up to 88\%—of its seen-class accuracy on the unseen split. These results demonstrate that even with limited supervision, the projection layer learns a effective general mapping between vision and language. Then, we perform dataset ablation to further investigate which aspects of the dataset contribute to enhancing generalization.

To understand how this generalization is possible, we further conduct a mechanistic interpretability analysis of the transformer layers of the frozen LLM. By extending techniques originally developed for analyzing LLMs, we find that the feed-forward network (FFN) behaves like a key-value memory system, processing both seen and unseen visual tokens in a structurally similar fashion.




Our contributions are three-fold:
\begin{itemize}
\item We propose a novel evaluation framework to assess the generalization ability of projection layers to unseen labels, and through rigorously designed experiments, we found that they robustly exhibit a non-trivial level of generalization to unseen labels.

\item We extend a mechanistic interpretability technique originally developed for LLMs to VLMs, analyzing how FFNs operate when processing visual tokens. From this analysis, we observe that although FFNs were trained exclusively on textual tokens, they are able to process visual tokens, and the underlying mechanism closely resembles that used for textual tokens.

\item Through systematic dataset ablation, we identify class label diversity as a key factor in enhancing generalization to unseen labels.
\end{itemize}
\section{Related Work}

\subsection{Connection Modules of VLMs}

VLMs consist of three modules: a vision encoder, a connection module, and an LLM and connection modules can be categorized into projection-based, query-based, and fusion-based approaches~\cite{yin2023survey}. Query-based methods use a fixed set of learnable query embeddings to cross-attend to visual features within Transformer layers. Flamingo~\cite{alayrac2022flamingo} is a representative example of the fusion-based approach. This design injects the vision encoder’s outputs into intermediate layers of the LLM via a cross-attention mechanism. While fusion-based methods allow flexible architecture and enable deep interaction between the vision encoder and the LLM, their high computational cost has made them less preferred in recent developments. In contrast, projection-based methods are lightweight and have gained popularity—even among recent state-of-the-art models~\cite{bai2025qwen2, li2024llava, chen2024expanding}—due to their simplicity and strong performance.

\subsection{Embedding Space of Vision and Text}
The Platonic Representation Hypothesis~\cite{huh2024platonic} posits the existence of an underlying representation which is independent of modality and argues that as models scale and solve more tasks, their embedding spaces naturally converge. 
They demonstrate this convergence across a range of vision models trained with different objectives and various language models. 
LiMBeR~\cite{MerulloCEP23} connects frozen vision and language models using a simple linear layer, showing that even vision models trained without explicit text supervision can be effectively linked to language, resulting in a VLM with strong performance. 
Together, these studies raise the question of whether a linear layer can also align unseen concepts, which motivates our proposed evaluation framework.

\subsection{Mechanistic Interpretability}

Mechanistic interpretability is a field of study that aims to uncover the internal processes of a model in order to understand how specific outputs are generated~\cite{liu2025mechanistic}. In recent years, language models have grown significantly in scale and capability, prompting increased attention to mechanistic interpretability as a means of improving model understanding, explainability, robustness, and transparency. Rather than being confined to a single technique, mechanistic interpretability encompasses a range of methodologies. For instance, the logit lens~\cite{nostalgebraist2020logitlens} examines how representations evolve across layers by projecting hidden states into the vocabulary space. Similarly, FFN key-value memory analysis~\cite{GevaSBL21} investigates the internal mechanisms of feed-forward networks.

More recently, methods originally designed for analyzing LLMs have been extended to vision-language models (VLMs). For example, \citet{NeoO0G0B25} applied the logit lens to VLMs and found that visual tokens become increasingly interpretable in the vocabulary space as they pass through the layers. In our work, we adapt FFN key-value memory analysis to VLMs and observe consistent findings with prior studies~\cite{NeoO0G0B25, GevaSBL21}.
\section{Evaluation Protocol for Unseen Label Generalization}

In this section, we present our method for converting object detection datasets into VLM-style prompts, with explicit control to ensure a clear separation between seen and unseen labels.

\subsection{Motivation and Design of the Dataset}
Consider the following VQA-style example for illustration:
\begin{quote}
\textbf{Q:} What is in the image?\
\textbf{A:} Grape
\end{quote}

The critical point in this example is that the word ``Grape’’ must not appear in the textual portion of the training data, as the training signal is derived from image-conditioned text generation. 

Since no publicly available dataset satisfies this constraint, we construct a synthetic dataset using object detection annotations. These annotations provide explicit class labels and corresponding regions, enabling us to partition the class labels into disjoint sets—seen and unseen—with a clear separation of visual concepts. Each bounding box annotation and its associated class label is then converted into a prompt of the following format:  
\begin{center}
\resizebox{\linewidth}{!}{\texttt{[image] bbox:[[x1],[y1],[x2],[y2]] [class label] \textbackslash n}}
\end{center} Using these prompts, we train the VLM to predict the class label conditioned on the image and bounding box coordinates. After training, we evaluate unseen-label generalization by constructing a multiple-choice QA benchmark separately for the seen and unseen label sets. By comparing performance on the unseen-class benchmark to that on the seen-class benchmark, we can quantify the model’s ability to generalize beyond the training labels.

\subsection{Dataset Curation}

We leverage Visual Genome~\cite{krishna2017visual} version 1.4 as our object detection dataset. We choose Visual Genome due to its extensive class diversity and fine-grained class labels, containing over 7,600 unique class labels—significantly more than other widely used object detection datasets such as COCO~\cite{lin2014microsoft} (80 classes) and Pascal VOC~\cite{everingham2010pascal} (20 classes). However, due to the long-tailed distribution of class frequencies, many rare classes are overly specific. To address this, we filter out classes with fewer than 10 instances.

Class labels in Visual Genome are provided in the format of WordNet~\cite{miller1995wordnet} synsets. For synsets that share the same lemma, we retain only the most frequent one and use its lemma as the class label.

To eliminate excessively small or large bounding boxes, we discard annotations whose bounding box area is smaller than 0.2\% or larger than 50\% of the image area.

We divide the remaining unique class labels into mutually exclusive seen and unseen groups. To minimize semantic overlap between the two, we embed all class labels using the all-MiniLM-L6-v2 model and perform spherical k-means clustering with $K=2$ on the resulting embeddings. This results in 1,780 classes in the seen group and 1,246 classes in the unseen group. Throughout this work, we consistently use all-MiniLM-L6-v2 whenever text embeddings are required in our experiments.

The dataset consists of 108{,}249 images, which we randomly split into 80\% for training and 20\% for testing, ensuring no image overlap. Each image is tagged as either train or test, and each class label is marked as seen or unseen. The model is trained using the train-seen subset and evaluated on both test-seen and test-unseen. In total, our split yields 701{,}777 prompts in train-seen, 176{,}135 prompts in test-seen, and 138{,}941 prompts in test-unseen.

\subsection{Evaluation Dataset and Protocol}

We conduct evaluation using a multiple-choice QA setting. Given a bounding box annotation, we randomly sample three distractor class labels from the same label group (seen or unseen) and construct a 4-choice QA. Following the protocol of SeedBench~\cite{li2024seed}, we compute the loss for each choice in the same manner as during training and consider the option with the lowest loss as the model's final prediction. This approach avoids unnecessary parsing and simplifies evaluation.

To address class imbalance, we exclude classes with fewer than 20 instances and randomly sample up to 200 instances for classes with more than 200. After this filtering, the test-seen-MCQA set contains 600 unique class labels and 61{,}272 QA examples, while the test-unseen-MCQA set includes 468 unique class labels and 48{,}724 QA examples. Despite the above measures, class imbalance remains due to many labels still having fewer than 200 instances. To mitigate this, we adopt a carefully chosen evaluation metric. Computing overall average accuracy would be biased toward dominant classes, so we instead report macro-averaged accuracy, which is calculated by first averaging accuracy per class and then computing the mean across all classes. For simplicity, we refer to macro-averaged accuracy as accuracy in the experiment section.

\subsection{Model Architecture}

Our model architecture follows that of LLaVA~\cite{liu2023visual}. It consists of a ViT-based vision encoder, a linear projection layer, and an LLM. Only the projection layer is trainable, while the other components remain fixed. Unless otherwise specified, we use clip-vit-large-patch14-336 as the vision encoder and Llama-3.2-3B-Instruct as the LLM as our default configuration. For reproducibility, we use model weights provided by the Hugging Face repositories. Following LLaVA, we exclude the [CLS] embedding from the vision encoder’s output and use only the patch tokens as input to the LLM.

\subsection{Prompt Design and Training Objective}

The goal of training is to predict the class label given the image and bounding box coordinates. To this end, the prompt is formatted as follows: \begin{center}
\resizebox{\linewidth}{!}{\texttt{[image] bbox:[[x1],[y1],[x2],[y2]] [class label] \textbackslash n}}\end{center} Here, \texttt{[image]} is a placeholder for the visual features, and \texttt{[BOS]} denotes the model-dependent beginning-of-sentence token. The box coordinates are rounded to the second decimal place. \texttt{\textbackslash n} serves as a model-agnostic end-of-sentence (EOS) token. We adopt this simple, model-agnostic prompt format rather than a model-specific chat template. This design follows the practice used in LLaVA's official implementation~\cite{llava2023github}. We compute the loss only on the tokens corresponding to the class label and the model-agnostic EOS token. 

\subsection{Implementation Details}

We use the AdamW optimizer and train the model for one epoch. The learning rate is set to 1e-3 without weight decay. The learning rate schedule includes a 3\% warm-up phase followed by cosine annealing. We use a batch size of 16, which fits within the GPU memory budget. All experiments are conducted on four NVIDIA A6000 GPUs. The loss function is the standard causal language modeling loss. Except for the batch size, all hyperparameters follow the official LLaVA pretraining configuration~\cite{llava2023github}, and we do not explore alternative values.
\section{Experiments}

The experiments are conducted in three stages, each with a distinct objective. The goal of the first stage is to validate whether unseen label generalization is consistently observed across various settings. Once this is confirmed, the second stage focuses on a dataset ablation study to identify which aspects of the dataset are important for enhancing unseen label generalization. Finally, in the third stage, we examine unseen label generalization through the lens of mechanistic interpretability. 

\subsection{Robustness of Unseen Label Generalization}
In this stage of experiments, we investigate whether unseen label generalization is consistently achieved across various settings. To this end, we design three experiments. Two are dedicated to model ablation: one varying the vision encoder and the other varying the LLM. Lastly, we construct a multiple-choice QA evaluation set based on OpenImages~\cite{krasin2017openimages}, which is entirely independent of the training set. Across all three rigorously designed experiments, we consistently find that projection layers achieve meaningful generalization toward unseen labels.

\subsubsection{Variation Across Vision Encoders}



Motivated by the Platonic Representation Hypothesis~\cite{huh2024platonic}, we select four representative vision encoders: DINOv2~\cite{oquab2023dinov2} and MAE~\cite{he2022masked} as self-supervised models, ViT~\cite{dosovitskiy2020image} trained on classification, and CLIP~\cite{radford2021learning}, thereby covering diverse training objectives. All models have a comparable number of parameters and are based on the ViT-Large architecture, which has approximately 300 million parameters. The corresponding model tags are dinov2-large, webssl-mae300m-ful1l2b-224, vit-large-patch16-224-in21k, and clip-vit-large-patch14-336, with pretrained weights obtained from the Hugging Face repositories.

We report accuracy on test-seen and test-unseen, in addition to accuracy, we also report two additional metrics: Relative Performance (Rel. Perf., defined as unseen accuracy divided by seen accuracy) and Relative Gain over Random (RGR). RGR quantifies the performance improvement over random guessing, which corresponds to 25\% accuracy in our 4-choice setting (e.g., RGR is 100\% when accuracy is 50\%). Table~\ref{tab:vision_encoder_ex1} presents the results from the vision encoder ablation study. All four encoders achieved RGR values exceeding 170\%, and Rel. Perf. scores ranged between 86\% and 89\%, indicating that unseen-label accuracy approaches upto 88\% of seen-label accuracy and is approximately 2.7 times higher than random guessing. These results suggest that vision encoders—even those trained without text-based supervision—can support non-trivial generalization to unseen labels. Nonetheless, CLIP, trained with text supervision, achieved the highest absolute accuracy, followed by DINOv2, the classification-trained ViT, and MAE.

Our findings exhibit a consistent pattern with the results reported in the Platonic Representation Hypothesis~\cite{huh2024platonic}. That work evaluated the alignment score between vision encoders and LLMs using a cross-modal dataset~\cite{srinivasan2021wit}, and showed that models trained without text supervision (e.g., DINOv2, MAE) yield vision embeddings that preserve a local neighborhood structure similar to that of LLMs. Their alignment scores, however, varied across models: highest for CLIP, followed by DINOv2, classification-trained ViT, and MAE. Strikingly, this ordering coincides with the ranking observed in our experiment (see the RGR scores in Table~\ref{tab:vision_encoder_ex1}). 

Despite substantial differences in experimental setup, the consistency between the two findings is non-trivial and reinforces confidence in the robustness of our results. Furthermore, we observe that a linear transformation can effectively transfer visual information from vision encoders to the LLM, even for unseen concepts. However, the relative effectiveness differs across models in a manner consistent with the prior study. Finally, we highlight that the purpose of this experiment is not to identify the best vision encoder, but to evaluate whether encoders trained with diverse objectives can support unseen-label generalization when integrated into a VLM.

\begin{table}[]
\centering
\begin{tabular}{ccccc}
\hline
       & \multicolumn{1}{l}{\begin{tabular}[c]{@{}c@{}}test-\\ seen (\%)$\uparrow$\end{tabular}} & \multicolumn{1}{l}{\begin{tabular}[c]{@{}c@{}}test-\\ unseen (\%)$\uparrow$\end{tabular}} & \multicolumn{1}{l}{\begin{tabular}[c]{@{}c@{}}Rel.\\  Perf. (\%)$\uparrow$\end{tabular}} & \multicolumn{1}{c}{RGR (\%)$\uparrow$} \\ \hline
MAE    & 77.3                                                                    & 66.9                                                                      & 86.5                                                                          & 167.6                        \\
ViT-cls    & 78.8                                                                    & 68.9                                                                      & 87.4                                                                          & 175.6                        \\
DinoV2 & 84.2                                                                    & 72.7                                                                      & 86.3                                                                          & 190.8                        \\
CLIP   & 84.2                                                           & 74.2                                                             & 88.1                                                                 & 196.8               \\ \hline
\end{tabular}
\caption{Results of the vision encoder ablation study. DINOv2~\cite{oquab2023dinov2} and MAE~\cite{he2022masked} are self-supervised models; ViT-cls~\cite{dosovitskiy2020image} is trained with classification supervision; and CLIP~\cite{radford2021learning} represents a text-supervised model. Accuracy is reported on test-seen and test-unseen. Rel. Perf. denotes accuracy on unseen labels relative to seen labels, and RGR indicates the relative gain over random guessing.}
\label{tab:vision_encoder_ex1}
\end{table}

\subsubsection{Variation Across LLMs}

We consider three configurations for the LLM ablation study: (1) smaller model size, (2) LLMs other than LLaMA~\cite{dubey2024llama}, and (3) pretrain-only models. Based on these criteria, we select Qwen3-0.6B, Qwen3-1.7B~\cite{yang2025qwen3}, and Llama-3.2-3B (pretrain-only). The results are presented in Table\ref{tab:llm_ex2}.

The pretrain-only model Llama-3.2-3B performs comparably to its instruction-tuned counterpart. Both achieve approximately 88\% Rel. Perf. and 195\% RGR. These results suggest that the ability to generalize to unseen labels may already be present from pretraining.

Even the smallest model, Qwen3-0.6B, achieves an RGR of 137.6\%, while Qwen3-1.7B attains an even higher value of 170.4\%. In terms of Rel. Perf., Qwen3-0.6B reaches 79.3\%, which is 5.8 percentage points lower than Qwen3-1.7B. Nonetheless, most configurations exhibit Rel. Perf. approximately 80\% or higher, indicating consistent unseen label generalization across different LLMs.

\begin{table}[]
\centering
\begin{tabular}{ccccc}
\hline
       & \multicolumn{1}{l}{\begin{tabular}[c]{@{}c@{}}test-\\ seen (\%)$\uparrow$\end{tabular}} & \multicolumn{1}{l}{\begin{tabular}[c]{@{}c@{}}test-\\ unseen (\%)$\uparrow$\end{tabular}} & \multicolumn{1}{l}{\begin{tabular}[c]{@{}c@{}}Rel.\\  Perf. (\%)$\uparrow$\end{tabular}} & \multicolumn{1}{l}{RGR (\%)$\uparrow$} \\ \hline             
\textit{Default}   & 84.2                          & 74.2                            & 88.1                           & 196.8                        \\
\textit{Pretrain}  & 84.1                          & 74.3                            & 88.3                           & 197.2                        \\
Q3-1.7B & 79.4                          & 67.6                            & 85.1                           & 170.4                        \\
Q3-0.6B & 74.9                          & 59.4                            & 79.3                           & 137.6                        \\ \hline
\end{tabular}
\caption{LLM ablation experiments are conducted on three additional LLMs with varying configurations. \textit{Default} refers to the LLaMA 3.2 3B~\cite{dubey2024llama} instruction-tuned model, and \textit{Pretrain} indicates its pretraining-only counterpart. Q3-1.7B~\cite{yang2025qwen3} and Q3-0.6B denote Qwen3-1.7B and Qwen3-0.6B, respectively. Metrics are reported using the same format as in Table~\ref{tab:vision_encoder_ex1}. Across various LLM configurations, even in the challenging case of a very small model (0.6B), we observe an RGR of 137.6, substantially exceeding random guessing and demonstrating a non-trivial degree of generalization.
}
\label{tab:llm_ex2}
\end{table}

\subsubsection{Evaluation on Out-of-Distribution Dataset (OpenImages)}

\citet{liu2024decade} points out that certain biases present in widely used datasets are not salient to humans but can be easily exploited by neural networks. To establish a more challenging evaluation environment, we perform experiments on OpenImages-V6, which is entirely distinct from Visual Genome. Moreover, the two datasets differ saliently in their class label styles in ways that are immediately apparent to humans. For example, OpenImages contains labels such as “Nail (Construction),” “Bronze sculpture,” “Lavender (Plant),” “Brown bear,” and “Rays and skates,” while Visual Genome uses much simpler object-centric terms such as “lamp,” “shelf,” “countertop,” “chain,” and “barroom.”

Following the same procedure used for constructing MCQA sets from Visual Genome, we omit classes with fewer than 20 instances and subsample up to 200 instances for classes with higher frequency. To remove identical or highly similar class labels, we leverage text embeddings of class names and filter out those with cosine similarity greater than 0.9. For each bounding box annotation, we construct a 4-choices QA item by randomly sampling three distractors from the OpenImages class pool. 

After preprocessing, we obtained 269 unique class labels and 27{,}374 QA samples. As shown in table~\ref{tab:openimages}, the VLM achieves an RGR of 174\% on the OpenImages MCQA set, performing 2.7 times better than random chance. Additionally, its accuracy on OpenImages reaches 92.5\% of that on test-unseen, indicating strong consistency. Despite differences in source images and annotation protocols between OpenImages and Visual Genome, 
the projection layer demonstrates robust generalization toward unseen labels.

\begin{table}[]
\centering
\begin{tabular}{cccc}
\hline
           & \begin{tabular}[c]{@{}c@{}}test-\\ seen (\%)$\uparrow$\end{tabular} & \multicolumn{1}{l}{\begin{tabular}[c]{@{}c@{}}test-\\ unseen (\%)$\uparrow$\end{tabular}} & \multicolumn{1}{l}{RGR (\%)$\uparrow$} \\ \hline
OpenImages & N/A                                                  & 68.7                                                                       & 174.8                        \\
VG         & \multicolumn{1}{c}{84.2}                             & 74.2                                                                       & 196.8                        \\ \hline
\end{tabular}
\caption{Results on OpenImages~\cite{krasin2017openimages} and Visual Genome~\cite{krishna2017visual}. Since the test set derived from OpenImages naturally falls into the test-unseen category, no score is reported for test-seen. Despite OpenImages being entirely independent, with class labels defined in a different style, the model still demonstrates solid performance comparable to that on VG.}

\label{tab:openimages}
\end{table}

\begin{figure}[t]
  \centering
  \includegraphics[width=\linewidth]{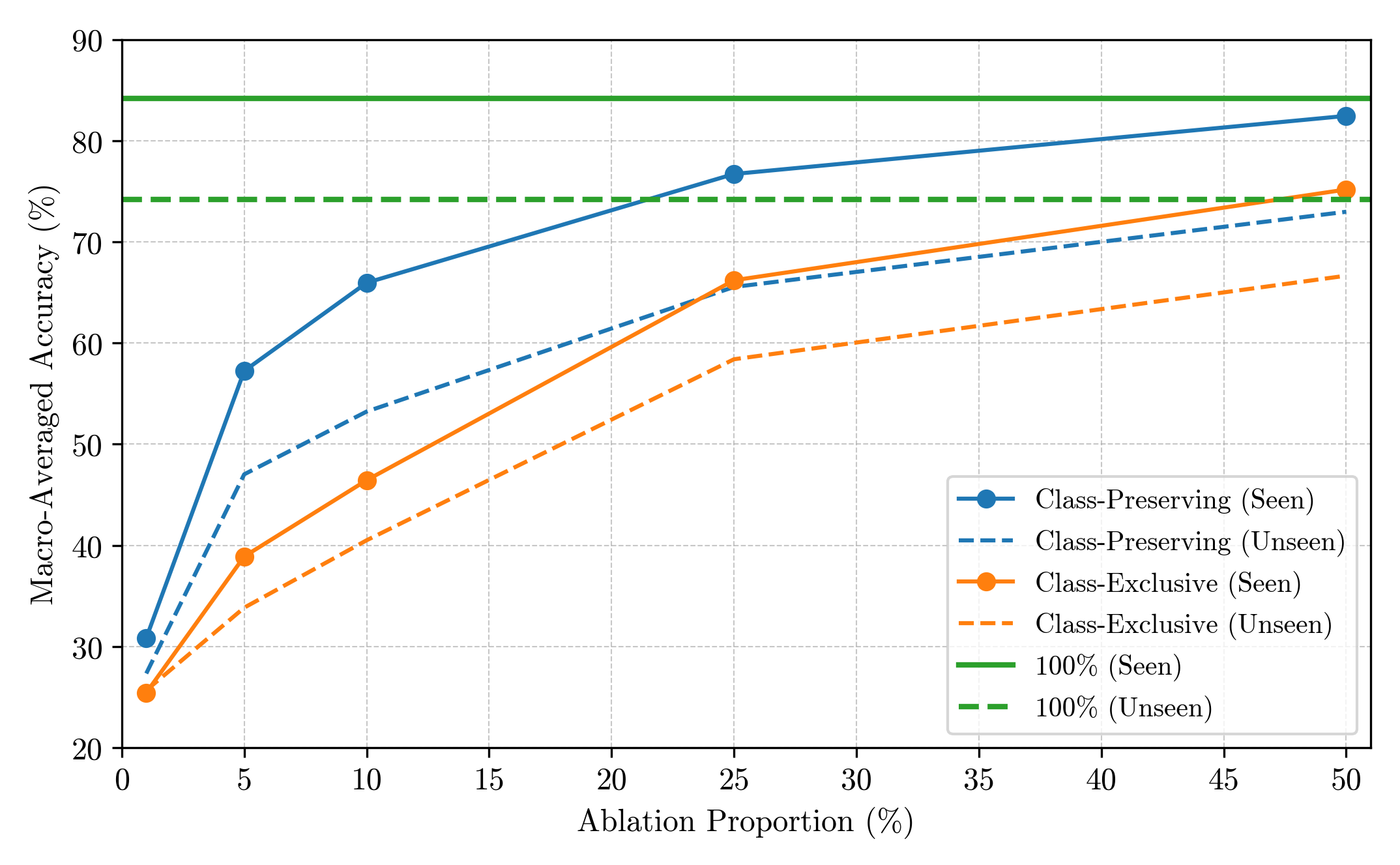}
  \caption{We present results from the dataset ablation study. Blue lines denote class-preserving methods and orange lines denote class-exclusive methods. Solid and dotted lines indicate test-seen and test-unseen performance, respectively. The x-axis shows the ablation proportion and the y-axis shows accuracy. Green horizontal lines represent results with the full (100\%) dataset. Across all proportions, class-preserving methods outperform class-exclusive methods, highlighting the importance of class diversity.}
  \label{fig:ablation}
\end{figure}

\subsection{Effect of Dataset Diversity}

In this experiment, we further conduct a dataset ablation study to identify which aspects are particularly beneficial for enhancing unseen label generalization. Specifically, we investigate whether class diversity contributes to improved generalization.

We reduce the dataset using two distinct methods. The first, class-exclusive, retains a certain proportion of unique class labels, with smaller subsets nested within larger ones. The second, class-preserving, keeps all unique class labels while downsampling instances uniformly across classes to match the total size of the class-exclusive counterpart.

We experiment with proportions of 50\%, 25\%, 10\%, 5\%, and 1\% for both ablation strategies. For each ablated dataset, the projection layer is trained from scratch, and evaluation is performed in exactly the same way as before. Figure~\ref{fig:ablation} shows that the class-preserving method consistently outperforms the class-exclusive method across all proportions. Notably, at 50\%, the performance drop in the class-preserving setting is minimal. These results suggest that class diversity plays a significant role in enhancing unseen label generalization.


\subsection{Investigating FFN on unseen label generalization}

We apply a mechanistic interpretability framework to examine how visual features are processed within the frozen LLM of a vision-language model, aiming to better understand unseen label generalization. In particular, we analyze the intermediate layers to study how projected visual tokens are handled by the LLM. These tokens are fundamentally different from text tokens, as they vary smoothly with changes in the input image and cover a dense region of the embedding space. By uncovering these processing mechanisms, we can further compare the treatment of visual tokens associated with seen and unseen labels, assessing whether they are processed in similar or distinct ways.


From this analysis, we arrive at three key findings: (1) FFNs function as (unnormalized) key–value memories when processing visual tokens, mirroring their role with textual tokens despite having been trained exclusively on text. The keys capture semantic patterns in the visual input, while the corresponding values store predictive information about the next token; (2) in lower layers, the embedding space in which the FFNs operate differs from the output embedding space, but these spaces become increasingly aligned in the upper layers; and (3) the mechanism for processing visual tokens is consistent across both seen and unseen labels. Each of these findings is consistent with prior work~\cite{NeoO0G0B25, GevaSBL21}.




\subsubsection{Brief Background on FFN Key-Value Interpretation}

The feed-forward network (FFN) in each Transformer block is a two-layer MLP composed of an expansion layer, a non-linear activation, and a compression layer. The expansion layer projects each input token $x \in \mathbb{R}^d$ to a higher-dimensional hidden space $\mathbb{R}^{d_m}$ (with $d_m > d$ in most cases). We denote the expansion layer’s weight matrix as $K \in \mathbb{R}^{d_m \times d}$. The activation output is $h = f(Kx),$
where $f$ is the non-linear activation function and $h \in \mathbb{R}^{d_m}$ represents the \emph{memory coefficients}. Each row vector of $K$, denoted $k_i^l$ for the $i$-th row in layer $l$, acts as a \emph{key vector}. The dot product $k_i^l \cdot x$ measures the similarity between the input token and the key, producing the corresponding memory coefficient. The compression layer is parameterized by $V \in \mathbb{R}^{d \times d_m}$, whose columns $v_j$ are \emph{value vectors} in $\mathbb{R}^d$. The final FFN output is a linear combination of value vectors weighted by the memory coefficients in $h$.

From this perspective, the FFN acts as an \emph{unnormalized key--value memory}. 
The keys effectively capture human-interpretable patterns in the input prefixes, 
and the corresponding values, when projected into the vocabulary space, 
concentrate probability mass on tokens that are likely to appear after those patterns.

\subsubsection{Constructing Visual Prefixes and Extracting Key-Value Pairs}


We hypothesize that the FFNs also act as key–value memories when processing visual tokens. To test this, we reformulate the prompts in the test-seen and test-unseen sets so that we can perform statistical analysis without manually inspecting each image. Specifically, we construct visual prefixes by removing the class label and the model-agnostic EOS token from each prompt. This format has a clear advantage: its semantics correspond to the image region confined by the bounding box, for which the ground-truth class is known a priori, and the next token following the prefix is deterministically the class label. On average, these prefixes are around 600 tokens long, of which approximately 576 are visual tokens, showing that they are overwhelmingly dominated by visual information. These properties allow us to analyze whether keys and values encode meaningful semantics without the need for manual inspection.

Using these prefixes, we extract key--value pairs from the model. For each prefix, we run a forward pass and identify the top three keys in each layer with the highest memory coefficients (i.e., the most activated keys) and their corresponding values. In our default LLM, there are $28$ layers and a hidden dimension of $8192$, meaning each layer contains $8192$ key--value pairs. This procedure yields three representative key--value pairs per layer for further analysis.

\subsubsection{Keys Encode Human-Interpretable Visual Semantics}

\begin{figure}[t]
  \centering
  \includegraphics[width=\linewidth]{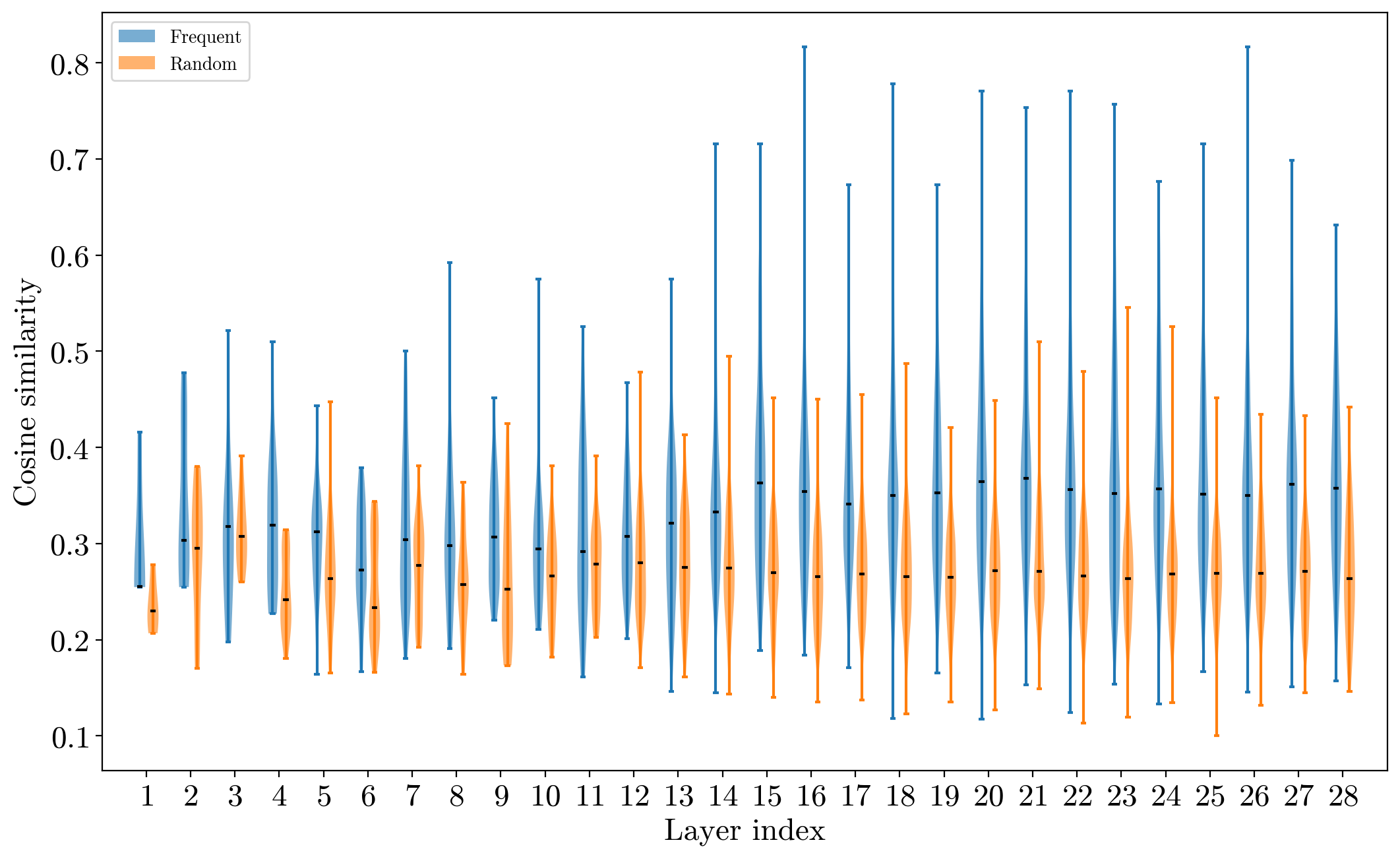}
  \caption{ 
  This shows violin plots of cosine similarity distributions from test-seen visual prefixes. For each key, we take the three most frequent class labels, compute pairwise cosine similarities among them, and plot the distributions per layer. Frequent class labels are shown in blue and the random baseline in orange, with black dots indicating distribution means.
  }
  \label{fig:key_analysis_seen}
\end{figure}
\begin{figure}[t]
  \centering
  \includegraphics[width=\linewidth]{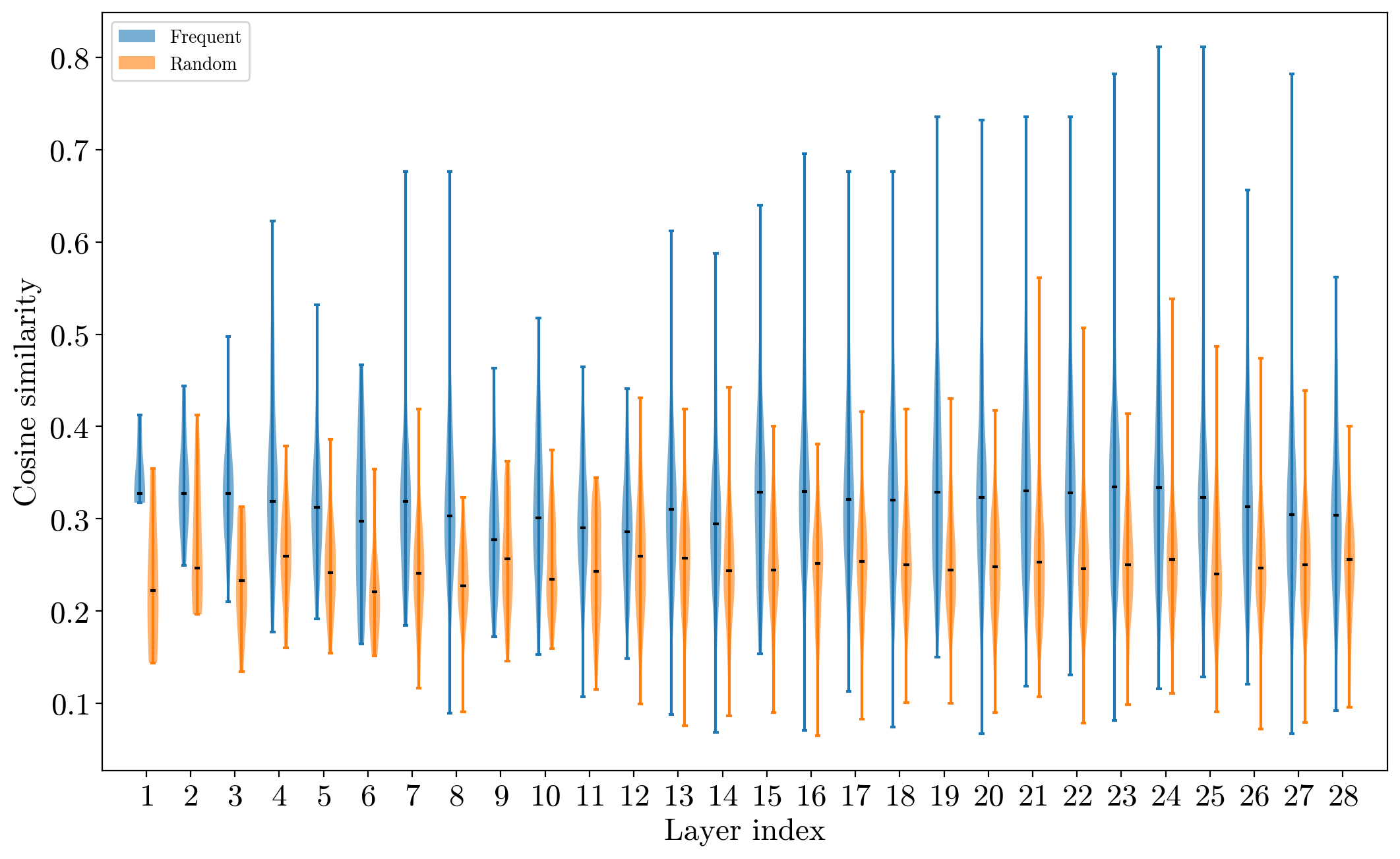}
  \caption{Layer-wise violin plots of cosine similarities for visual prefixes in the test-unseen set. Notations, colors, and axis definitions are the same as in Figure~\ref{fig:key_analysis_seen}.}
  \label{fig:key_analysis_unseen}
\end{figure}

We obtain 6,359 and 5,744 unique keys from the visual prefixes in the test-seen and test-unseen sets, respectively. To examine whether the visual prefixes associated with a given key are semantically clustered, we collect the class labels of all visual prefixes that activate that key. For each key, we then retain the three most frequent class labels. As a random baseline, we emulate this process by randomly sampling three class labels from within the same label group. Finally, for each set of three class labels (real or baseline), we extract text embeddings and compute the cosine similarity across all possible pairs. 
Figure~\ref{fig:key_analysis_seen} presents violin plots of cosine similarities across layers for the test-seen set, while Figure~\ref{fig:key_analysis_unseen} shows the corresponding plots for the test-unseen set. In both cases, the real class labels (blue) consistently exhibit higher cosine similarities than the random baseline (orange) across all layers. This pattern is also reflected in the averages (black dots), as well as in the maximum values, which extend notably higher in the upper layers. These results indicate that keys effectively capture semantics in the input prefixes, and this mechanism holds consistently for both seen and unseen labels.




\subsubsection{Investigating Projected Value at Layerwise Level}

\begin{figure}[h]
  \centering
  \includegraphics[width=\linewidth]{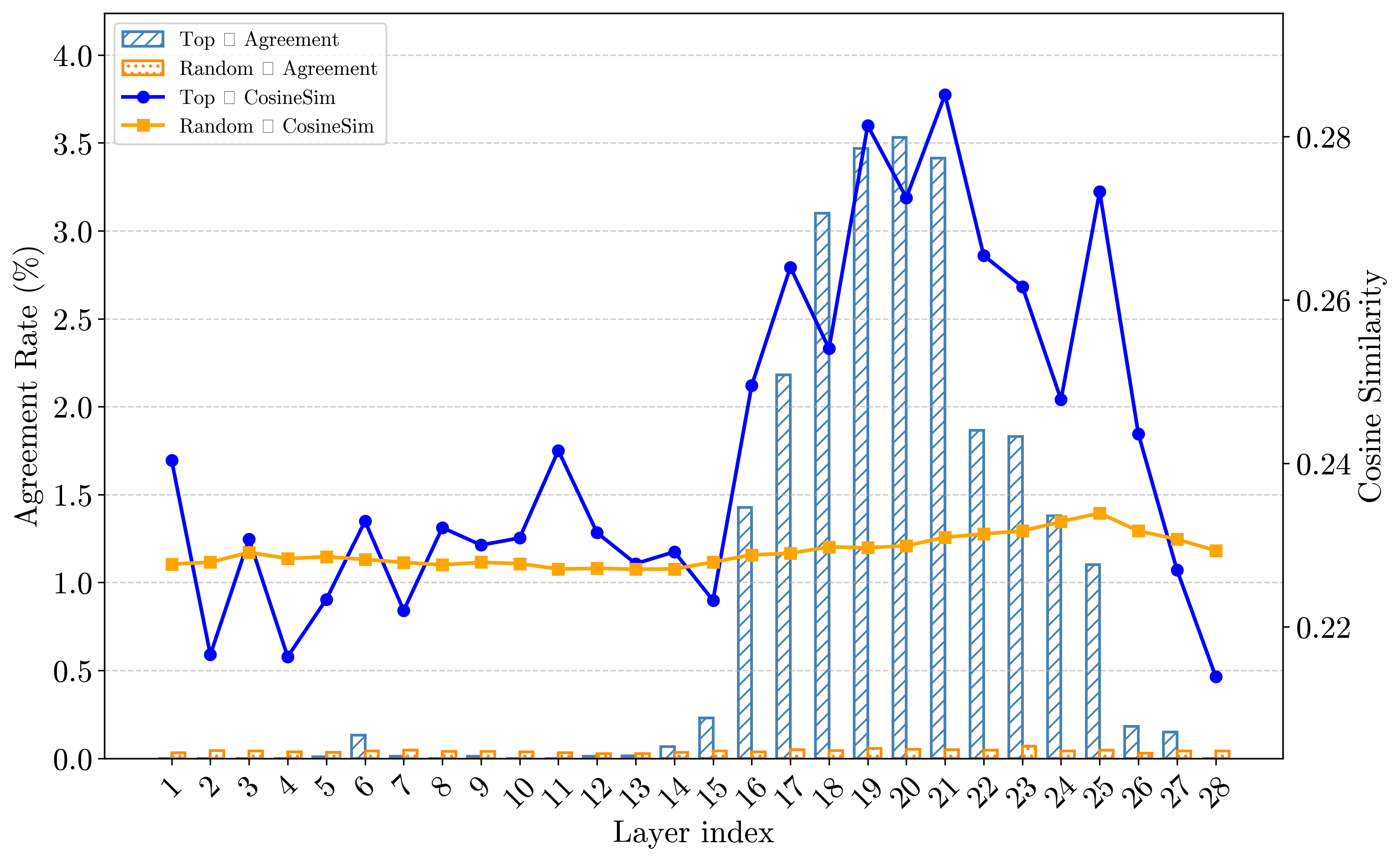}
  \caption{
  This plot shows agreement rate and cosine similarity obtained from test-seen set between tokens extracted from value vectors and the next token, layer by layer. Results are shown for values with the highest memory coefficients and for randomly sampled values (marked as Top and Random in the legend). The x-axis is the layer index. The left y-axis gives agreement rate (bar plots), and the right y-axis gives cosine similarity (line plots).}
  \label{fig:ffn_value_seen}
\end{figure}

\begin{figure}[h]
  \centering
  \includegraphics[width=\linewidth]{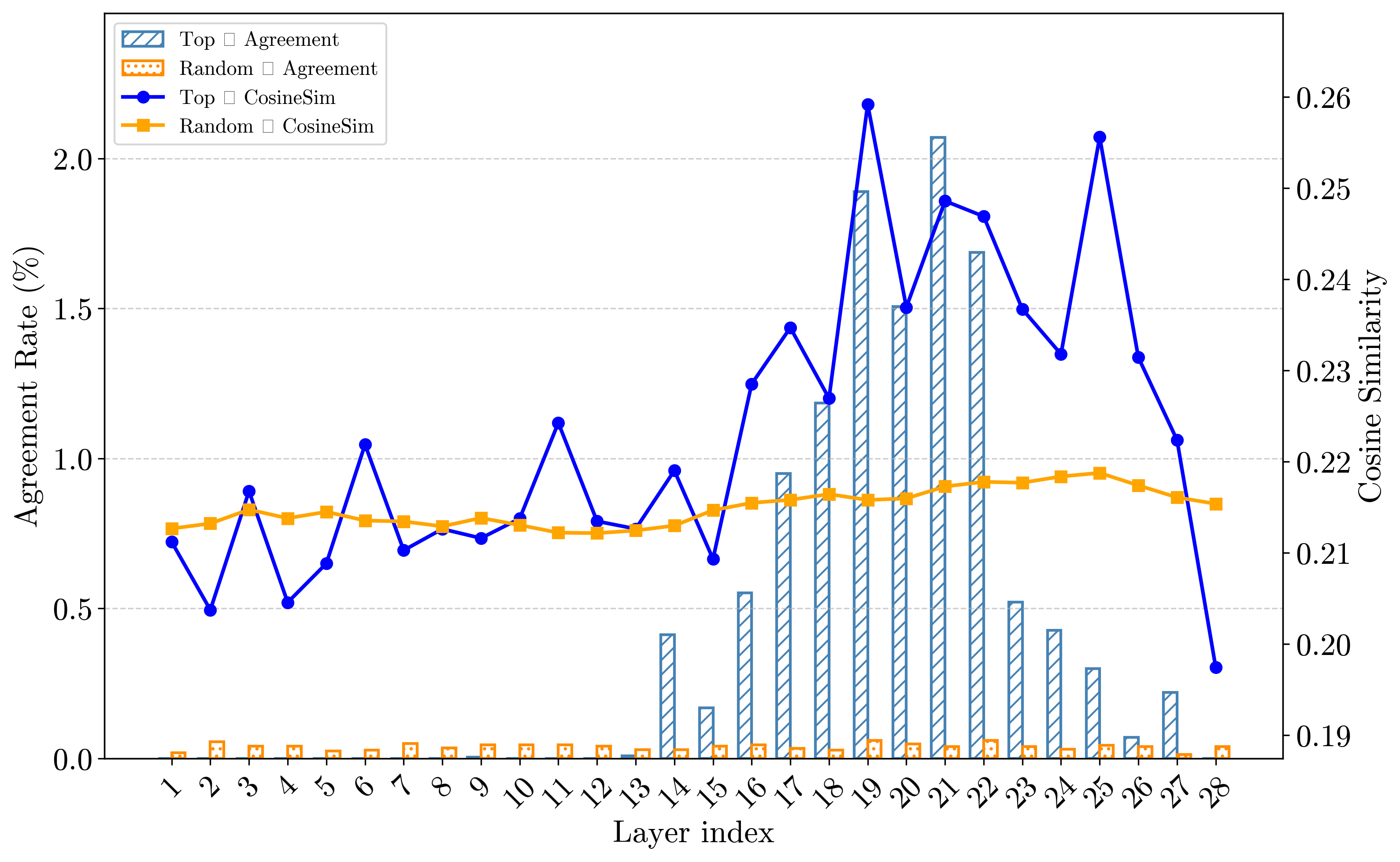}
  \caption{This plot shows the same metrics as in Fig.~\ref{fig:ffn_value_seen}, but obtained from the test-unseen set.}
  \label{fig:ffn_value_unseen}
\end{figure}


We investigate whether value vectors associated with high memory coefficients encode information about the next token. Specifically, we project the values into the vocabulary space by multiplying them with the output embedding matrix, yielding unnormalized logits. From each value vector, we then extract the top three tokens with the highest logit values and examine whether these tokens are semantically close to the actual next token. For comparison, we construct a random baseline by selecting three random values per layer, from which we likewise extract three tokens in the same manner. Following~\citet{GevaSBL21}, we report agreement rates between the extracted tokens and the ground-truth class label, and additionally measure cosine similarity between embeddings as a smoother metric beyond binary matches.

Figure~\ref{fig:ffn_value_seen} presents the layer-wise agreement rates and cosine similarities for the test-seen set. Both metrics remain near the random baseline in lower layers but rise notably after layer 15, eventually surpassing the baseline by a clear margin. Figure~\ref{fig:ffn_value_unseen} shows the same analysis for the test-unseen set. The curves exhibit nearly identical shapes to those of the test-seen set, indicating a consistent trend across both conditions.

These results yield several insights. First, value vectors associated with high memory coefficient contain distributional information about the next token, even when prefixes are dominated by visual inputs. Second, value vectors become increasingly interpretable in the vocabulary space as depth increases, a pattern observed for both seen and unseen labels. This suggests that FFNs operate through the same underlying mechanism regardless of label familiarity, although the degree of alignment may vary. The findings align with prior observations that value vectors become more interpretable in higher layers~\citep{GevaSBL21}, and they echo results from~\citet{NeoO0G0B25}, where visual tokens were shown to gain interpretability in later layers. Notably, both agreement rates and cosine similarity drop at the final layer, which we interpret as evidence that the last layer performs post-processing—such as suppressing noisy components—rather than directly estimating next-token probabilities~\citep{lad2024remarkable}.

\subsubsection{Qualitative Analysis of Key–Value Pairs}
Finally, we present qualitative examples of key–value pairs that satisfy two criteria: (1) the top three most frequent class labels associated with the key are semantically coherent, and (2) the corresponding value, when projected into the vocabulary space, ranks the class label among its top three logits. Table~\ref{tab:qualitative_keys} illustrates such examples: five from the test-seen set and five from the test-unseen set. For instance, the first row of Table~\ref{tab:qualitative_keys} shows an exemplar key–value pair from the test-seen set, $k_{2663}^{16}$, which is activated by visual prefixes whose class label is \textit{tooth} (42.9\%), followed by \textit{lip} (12.2\%) and \textit{smile} (10.2\%). These labels form a semantically cohesive group, all belonging to the human mouth region. Similar patterns are exhibited across the remaining examples from the test-seen set (rows 2–5). Rows 6–10 of Table~\ref{tab:qualitative_keys} show analogous results for the test-unseen set. For example, row 6 includes $k_{1135}^{16}$, which is activated by prefixes labeled \textit{beach} (46.7\%), \textit{shore} (14.5\%), and \textit{ocean} (11.3\%). These labels are closely tied to the concept of the sea, again demonstrating semantic coherence. Comparable trends are observed in other unseen examples, reinforcing that the behavior of key–value pairs generalizes beyond seen labels.

\begin{table}[!htbp]
\small                      
\setlength{\tabcolsep}{4pt}         
\renewcommand{\arraystretch}{0.9}   
\centering
\begin{tabular}{ccccc}
\hline
Keys                                                                                & \begin{tabular}[c]{@{}c@{}}Top tokens\\from value\end{tabular} & Top1                                                                           & Top2                                                                        & Top3                                                                             \\ \hline
\multirow{3}{*}{\begin{tabular}[c]{@{}c@{}}$k_{2663}^{16}$\\ - seen\end{tabular}}   & teeth                                                                      & \multirow{3}{*}{\begin{tabular}[c]{@{}c@{}}tooth\\ (42.9\%)\end{tabular}}      & \multirow{3}{*}{\begin{tabular}[c]{@{}c@{}}lip\\ (12.2\%)\end{tabular}}     & \multirow{3}{*}{\begin{tabular}[c]{@{}c@{}}smile\\ (10.2\%)\end{tabular}}        \\
                                                                                    & tooth                                                                      &                                                                                &                                                                             &                                                                                  \\
                                                                                    & dental                                                                     &                                                                                &                                                                             &                                                                                  \\ \hline
\multirow{3}{*}{\begin{tabular}[c]{@{}c@{}}$k_{7582}^{19}$\\ - seen\end{tabular}}   & cat                                                                        & \multirow{3}{*}{\begin{tabular}[c]{@{}c@{}}cat\\ (78.9\%)\end{tabular}}        & \multirow{3}{*}{\begin{tabular}[c]{@{}c@{}}fur\\ (8.8\%)\end{tabular}}      & \multirow{3}{*}{\begin{tabular}[c]{@{}c@{}}rug\\ (1.8\%)\end{tabular}}           \\
                                                                                    & Cat                                                                        &                                                                                &                                                                             &                                                                                  \\
                                                                                    & pur                                                                        &                                                                                &                                                                             &                                                                                  \\ \hline
\multirow{3}{*}{\begin{tabular}[c]{@{}c@{}}$k_{5606}^{19}$\\ - seen\end{tabular}}   & Dough                                                                      & \multirow{3}{*}{\begin{tabular}[c]{@{}c@{}}dough\\ (17.1\%)\end{tabular}}      & \multirow{3}{*}{\begin{tabular}[c]{@{}c@{}}bagel\\ (12.9\%)\end{tabular}}   & \multirow{3}{*}{\begin{tabular}[c]{@{}c@{}}doughnut\\ (8.6\%)\end{tabular}}      \\
                                                                                    & puff                                                                       &                                                                                &                                                                             &                                                                                  \\
                                                                                    & dough                                                                      &                                                                                &                                                                             &                                                                                  \\ \hline
\multirow{3}{*}{\begin{tabular}[c]{@{}c@{}}$k_{1331}^{17}$\\ - seen\end{tabular}}   & sheep                                                                      & \multirow{3}{*}{\begin{tabular}[c]{@{}c@{}}sheep\\ (47.4\%)\end{tabular}}      & \multirow{3}{*}{\begin{tabular}[c]{@{}c@{}}wool\\ (11.6\%)\end{tabular}}    & \multirow{3}{*}{\begin{tabular}[c]{@{}c@{}}goat\\ (10.5\%)\end{tabular}}         \\
                                                                                    & Sheep                                                                      &                                                                                &                                                                             &                                                                                  \\
                                                                                    & lamb                                                                       &                                                                                &                                                                             &                                                                                  \\ \hline
\multirow{3}{*}{\begin{tabular}[c]{@{}c@{}}$k_{4954}^{17}$\\ - seen\end{tabular}}   & glasses                                                                    & \multirow{3}{*}{\begin{tabular}[c]{@{}c@{}}sunglasses\\ (17.0\%)\end{tabular}} & \multirow{3}{*}{\begin{tabular}[c]{@{}c@{}}goggles\\ (12.9\%)\end{tabular}} & \multirow{3}{*}{\begin{tabular}[c]{@{}c@{}}hat\\ (7.3\%)\end{tabular}}           \\
                                                                                    & Glasses                                                                    &                                                                                &                                                                             &                                                                                  \\
                                                                                    & sunglasses                                                                 &                                                                                &                                                                             &                                                                                  \\ \hline
\multirow{3}{*}{\begin{tabular}[c]{@{}c@{}}$k_{1135}^{16}$\\ - unseen\end{tabular}} & beach                                                                      & \multirow{3}{*}{\begin{tabular}[c]{@{}c@{}}beach\\ (46.7\%)\end{tabular}}      & \multirow{3}{*}{\begin{tabular}[c]{@{}c@{}}shore\\ (14.5\%)\end{tabular}}   & \multirow{3}{*}{\begin{tabular}[c]{@{}c@{}}ocean\\ (11.3\%)\end{tabular}}        \\
                                                                                    & Beach                                                                      &                                                                                &                                                                             &                                                                                  \\
                                                                                    & beaches                                                                    &                                                                                &                                                                             &                                                                                  \\ \hline
\multirow{3}{*}{\begin{tabular}[c]{@{}c@{}}$k_{6768}^{16}$\\ - unseen\end{tabular}} & boat                                                                       & \multirow{3}{*}{\begin{tabular}[c]{@{}c@{}}boat\\ (34.3\%)\end{tabular}}       & \multirow{3}{*}{\begin{tabular}[c]{@{}c@{}}raft\\ (11.4\%)\end{tabular}}    & \multirow{3}{*}{\begin{tabular}[c]{@{}c@{}}motor\\ (8.6\%)\end{tabular}}         \\
                                                                                    & boats                                                                      &                                                                                &                                                                             &                                                                                  \\
                                                                                    & Boat                                                                       &                                                                                &                                                                             &                                                                                  \\ \hline
\multirow{3}{*}{\begin{tabular}[c]{@{}c@{}}$k_{6496}^{15}$\\ - unseen\end{tabular}} & labeled                                                                    & \multirow{3}{*}{\begin{tabular}[c]{@{}c@{}}cabinet\\ (17.6\%)\end{tabular}}    & \multirow{3}{*}{\begin{tabular}[c]{@{}c@{}}drawer\\ (15.3\%)\end{tabular}}  & \multirow{3}{*}{\begin{tabular}[c]{@{}c@{}}refrigerator\\ (11.8\%)\end{tabular}} \\
                                                                                    & cabinet                                                                    &                                                                                &                                                                             &                                                                                  \\
                                                                                    & drawers                                                                    &                                                                                &                                                                             &                                                                                  \\ \hline
\multirow{3}{*}{\begin{tabular}[c]{@{}c@{}}$k_{935}^{17}$\\ - unseen\end{tabular}}  & clock                                                                      & \multirow{3}{*}{\begin{tabular}[c]{@{}c@{}}clock\\ (27.9\%)\end{tabular}}      & \multirow{3}{*}{\begin{tabular}[c]{@{}c@{}}wrist\\ (11.5\%)\end{tabular}}   & \multirow{3}{*}{\begin{tabular}[c]{@{}c@{}}numeral\\ (9.4\%)\end{tabular}}       \\
                                                                                    & clocks                                                                     &                                                                                &                                                                             &                                                                                  \\
                                                                                    & watch                                                                      &                                                                                &                                                                             &                                                                                  \\ \hline
\multirow{3}{*}{\begin{tabular}[c]{@{}c@{}}$k_{2}^{17}$\\ - unseen\end{tabular}}    & table                                                                      & \multirow{3}{*}{\begin{tabular}[c]{@{}c@{}}table\\ (7.7\%)\end{tabular}}       & \multirow{3}{*}{\begin{tabular}[c]{@{}c@{}}tray\\ (7.4\%)\end{tabular}}     & \multirow{3}{*}{\begin{tabular}[c]{@{}c@{}}disk\\ (6.8\%)\end{tabular}}          \\
                                                                                    & Table                                                                      &                                                                                &                                                                             &                                                                                  \\
                                                                                    & tables                                                                     &                                                                                &                                                                             &                                                                                  \\ \hline
\end{tabular}
\caption{
Qualitative examples of key–value pairs. We present five examples from the test-seen set (marked with “-seen” in the Keys column) and five from the test-unseen set (marked with “-unseen”). The Keys column specifies the layer and index of each key. The second column lists the top three tokens obtained by projecting the corresponding value into the vocabulary space (top-3 logits). The next three columns show the top three most frequent class labels associated with each key, along with the proportion of visual prefixes assigned to each label among all prefixes that activated the key.}
\label{tab:qualitative_keys}
\end{table}
\section{Conclusion}

We investigated whether the projection layer exhibits generalization capability to unseen labels. To this end, we proposed a novel framework for evaluating unseen label generalization leveraging object detection datasets. Our findings show that projection layers indeed demonstrate generalization to unseen labels across a range of rigorously designed experimental settings. Furthermore, mechanistic interpretability analysis provided supporting evidence that the FFN processes vision tokens associated with both seen and unseen labels in a consistent manner.
These results offer empirical support for the current architectural design of VLMs, which connect two foundational modules via a lightweight projection layer. Our findings suggest that the representation spaces of the vision encoder and the LLM are well aligned, enabling generalization beyond the alignment training set.
One limitation of our study is the reliance on a fixed prompt format. Future work may extend this investigation to more general visual question-answering tasks.
\FloatBarrier

\bibliography{aaai2026}
\end{document}